\documentclass[runningheads]{llncs}
\usepackage[T1]{fontenc}
\usepackage{graphicx}
\usepackage{color}

\usepackage[toc,page]{appendix}

\usepackage{caption}
\usepackage{subcaption}
\usepackage{amsmath,graphicx,amssymb,mathabx}
\usepackage{hyperref}
\usepackage{enumitem}
\setlist{nosep, leftmargin=14pt}

\usepackage{mwe} 
\usepackage[ruled,vlined]{algorithm2e}
\usepackage[utf8]{inputenc}
\usepackage[most]{tcolorbox}
\usepackage{bbm}
\usepackage{theorem}
\usepackage[export]{adjustbox}
\usepackage{xcolor}
\usepackage{multicol}
\usepackage{makecell}

\newcommand{\discint}[2]{[\![ #1, #2 ]\!]}

\newcommand{\partsof}[1]{\mathcal{P}(#1)}

\newcommand{\card}[1]{|#1|}

\newcommand{\R}{\mathbb{R}}

\newcommand{\cC}{\mathcal{C}}
\newcommand{\N}{\mathbb{N}}
\newcommand{\Z}{\mathbb{Z}}

\newcommand{\I}{\mathcal{I}}

\newcommand{\dil}[2]{#1 \oplus #2}
\newcommand{\ero}[2]{#1 \ominus #2}
\newcommand{\clos}[2]{#1 \bullet #2}

\newcommand{\ope}[2]{#1 \circ #2}
\newcommand{\compl}[1]{#1^C}
\newcommand{\whitehat}[2]{T_w(#1, #2)}
\newcommand{\blackhat}[2]{T_b(#1, #2)}
\newcommand{\indicator}[1]{\mathbbm{1}_{#1}}
\newcommand{\conv}[2]{#1 \circledast #2}

\newcommand{\bise}{\epsilon}
\newcommand{\lui}{\ensuremath{\text{LUI}}}
\newcommand{\bisel}{\phi}
\newcommand{\bimonn}{\ensuremath{\text{BiMoNN}}}

\newcommand{\almostbinary}[2]{\I(#1, #2)}
\newcommand{\mnist}{MNIST}
\newcommand{\diskorect}{Diskorect}

\newcommand{\argmin}[1]{\underset{#1}{\text{argmin}}\text{ }}

\begin{document}
\title{Binary Multi Channel Morphological Neural Network}
%
%
\author{Theodore Aouad\inst{1} \and
Hugues Talbot\inst{1}}
\authorrunning{T. Aouad and H. Talbot}
%
\institute{CentraleSup\'elec, Universit\'e Paris-Saclay, Inria. Gif-sur-Yvette, France
\email{\{firstname.lastname\}@centralesupelec.fr}}
\maketitle              
\begin{abstract}

Neural networks and particularly Deep learning have been comparatively little studied from the theoretical point of view. Conversely, Mathematical Morphology is a discipline with solid theoretical foundations. We combine these domains to propose a new type of neural architecture that is theoretically more explainable. We introduce a Binary Morphological Neural Network (BiMoNN) built upon the convolutional neural network. We design it for learning morphological networks with binary inputs and outputs. We demonstrate an equivalence between BiMoNNs and morphological operators that we can use to binarize entire networks.
These can learn classical morphological operators and show promising results on a medical imaging application.

\end{abstract}
\section{Introduction}
\label{sec:intro}

While demonstrating considerable success in applications, there are few theoretical results in Deep Learning. Many elements are not well understood, with networks operating as black boxes, hindering critical applications such as medical imaging or robotics. Conversely, Mathematical Morphology (MM) is a computer vision discipline with solid theoretical foundations. In this article, we propose to combine ideas from the two domains to construct a new type of neural architecture that is theoretically more justified and explainable. As argued in~\cite{angulo2021some} combining the two fields is promising, as MM can se used to construct simpler and more understandable deep neural networks. Using MM is more natural than convolutional neural networks (ConvNets) for some specific tasks. For example, ConvNets are not designed to deal with binary images, while MM is constructed on set-representation of images. Also, as stated in \cite{kirszenberg2021going}, designing a correct sequence of morphological operators can be complicated and time-consuming. 

Past researchers have proposed to learn both operators and structuring elements, e.g using the max-plus definition of dilations and erosion \cite{mondal2020image,franchi2020deep}. Other approaches introduce differentiable approximations of the max and min operators, such as the adaptive morphological layer \cite{shen2019deep}, the PConv layer \cite{masci2013learning}, and the most recent $\mathcal{L} Morph$ and $\mathcal{S} Morph$ layers \cite{kirszenberg2021going}. However, these methods deal with grey-scale morphology and grey-scale images. 

We introduces the Binary Morphological Neural Network (BiMoNN), a learnable morphological network with binary inputs and output, inspired by the structure of ConvNets. We replace the convolution by elementary operations from the set of erosions, dilations, anti-erosions, and anti-dilations: we can learn the type of operation and its associated structuring element as well as any intersection/union of these elementary operations.
we show equivalence properties between our neural network and morphological operators. They can be used both as explainable results or as a way to binarize the network: once these equivalences are respected, the morphological operation can replace the neural network. Booleans weights can replace real-valued weights, improving the inference's computing efficiency \cite{hubara2016binarized,kim2016bitwise,simons2019review}. We demonstrate how we managed to learn some classical morphological operators as well as an intermediate step of a challenging medical imaging problem. A preliminary version of our work is available at \cite{aouad2022binary}. Our code is publicly available online at \url{https://github.com/TheodoreAouad/Bimonn_DGMM2022}.







\section{Method}
\label{sec:method}

\subsection{Notations}

Let $\N^*$ be the set of non zero natural integers. Let $d \in \N^*$ the dimension of the image (usually $d=2$ or $d=3$). If $a \leq b \in \Z$, we denote the discrete interval as $\discint{a}{b} = [a, b] \cap \Z$. Let $N_I \in \N^*$ and $\Omega_I = \discint{-N_I}{N_I}^d$ be the support of binary images $I \subset \Omega_I$. Let $\Omega_S = \discint{-N_S}{N_S}^d$ be the support of structuring elements  (short: SE) $S \subset \Omega_S$. If $X \subset \Z^d$, we denote $\check{X} = \{-x ~|~ x \in X\}$ the symmetric of $X$ with respect to the origin. For a binary image $I \subset \Omega_I$, we define its complementation $\compl{I} = \Omega_I ~\backslash~ I$. For a SE $S \subset \Omega_S$, if there is no ambiguity, we define its complementation as $\compl{S} = \Omega_S ~\backslash~ S$. For a set $X \subset \Z^d$, we denote its indicator function $\indicator{X}: \Z^d \mapsto \R$ such that $\indicator{X}(x) = 1$ if $x \in X$, else $\indicator{X}(x) = 0$. If $\indicator{I} \in \R^{\Omega_I}$ and $\indicator{S} \in \R^{\Omega_S}$ we denote the convolutional product $\conv{\indicator{I}}{\indicator{S}}: k \in \Omega_I \mapsto \sum_{i \in \Omega_S | k-i \in \Omega_I}(\indicator{I}(k -i)\indicator{S}(i)))$.

\begin{definition}
Let $S \subset \Omega_S$ be a SE. Let $I \subset \Omega_I$ be a binary image. We define the following morphological operators.

\begin{align}
    \text{the dilation by S  } ~~&\delta_S(I) = \dil{I}{S} = \bigcup_{s \in S}{(I + s)} \\
    \text{the erosion by S  } ~~&\varepsilon_S(I) = \ero{I}{\check{S}} = \bigcap_{s \in S}(I - s) \\
    \text{the anti-dilation by S  } ~~&\compl{\delta_S(I)}\\
    \text{the anti-erosion by S  } ~~&\compl{\varepsilon_S(I)}\\
    \text{the opening by S  } ~~&\ope{I}{S} = \dil{(\ero{I}{\check{S}})}{S} \\
    \text{the closing by S  } ~~&\clos{I}{S} = \ero{(\dil{I}{S})}{\check{S}} \\
    \text{the black top-hat by S  }  ~~&\blackhat{I}{S} = (\clos{I}{S}) ~\backslash~ I = (\clos{I}{S}) \cap \compl{I}  \\
    \text{the white top-hat by S  }  ~~&\whitehat{I}{S} = I ~\backslash~ (\ope{I}{S}) = I \cap \compl{(\ope{I}{S})}
\end{align}

\end{definition}

While not exhaustive, this list is representative of many useful morphological operators. These operators are obtained through composition and intersection or union of dilations, erosions, anti-dilations, and anti-erosions. Therefore, it is enough to learn these elementary operations, their SEs, and their aggregation.

\subsection{Binary Structuring Element Neuron}

First, we design a single neuron that replaces the convolution operation: it can learn the dilation, erosion, anti-dilation, anti-erosion, and their SEs.

\subsubsection{Definition}

\paragraph{}Proposition \ref{prop:conv-morp} shows that we can express a dilation and erosion exactly using a thresholded convolution.

\begin{proposition}[Morphological operators from convolution]\label{prop:conv-morp}
Let $S \subset \Omega_S$ be a binary SE and $X \subset \Omega_I$ be a binary image. 
  \begin{align}
  \dil{X}{S} &= \bigg(\conv{\indicator{X}}{\indicator{S}} \geq 1\bigg) = \bigg(\conv{\indicator{X}}{\indicator{S}} - 1 \geq 0\bigg)\\   
  \ero{X}{S} &= \bigg(\conv{\indicator{X}}{\indicator{S}} \geq \card{S}\bigg) = \bigg(\conv{\indicator{X}}{\indicator{S}} - \card{S} \geq 0\bigg)
  \end{align}

\end{proposition}

We stress that the dilations and erosions only differ by a scalar, $1$ for the dilation and $\card{S}$ for the erosion. Given $S$, we can learn the operation using only this scalar. Taking inspiration from these expressions, we relax $S$ into real weights $W \in \R^{\Omega_S}$. We denote the softplus function by $f^+: x \in \R \mapsto \ln(1 + \exp(x))$. Let $\xi : \R \mapsto [0, 1]$ be a smooth increasing function such that $\xi(x) \rightarrow_{x \rightarrow -\infty} 0$ and $\xi(x) \rightarrow_{x \rightarrow +\infty} 1$. In practice, $\xi(x) = \frac{1}{2}\tanh(x)+\frac{1}{2}$.

\begin{definition}[BiSE neuron]\label{def:bise}
Let $W \in \R^{\Omega_S}$ be a weight matrix, $b \in \R$ a bias and $p \in \R$ a scaling number. We define a \textbf{BiSE (Binary Structuring Element) neuron} as follow:
\begin{equation}
\bise^{W, b, p}: x \in [0, 1]^{\Z^d} \mapsto \xi(p [\conv{x}{f^+(W)} - f^+(b)]) \in [0, 1]^{\Z^d}
\end{equation}        
\end{definition}

This expression approximates the thresholded convolution of proposition \ref{prop:conv-morp}. Indeed, we apply the convolution and subtract a bias. If this expression is negative, we want the result to be $0$, else we want the result to be $1$. Therefore, we multiply it by a scaling factor and threshold it. The higher $|p|$ is, the closer the output is to $0$ or $1$. We enforce the weights and bias to be positive: we will explain that later.

The BiSE neuron is a convolution layer with the weights and bias forced to be positive, with a smooth thresholding function as activation. In practice, all the operations are differentiable with real inputs and outputs, not binary. Thus, we define almost binary images, with pixels value either close to $0$ or close to $1$.

\begin{definition}[Almost Binary Image] 
We say an image $I \in [0, 1]^{\Omega_S}$ is \textbf{almost binary} if there exists $u < v \in [0, 1]$ such that $I(\Omega_S) \cap ]u, v[ = \emptyset$. We denote this set $\almostbinary{u}{v}$.
\end{definition}

\subsubsection{Morphological equivalence}

\paragraph{}The following Theorem states the conditions ensuring that the BiSE neuron is equivalent to a morphological operator. We denote $\bise^{W, b, +\infty}: x \in [0, 1]^{\Omega_I} \mapsto (\conv{x}{W} > b) \in \Omega_I$ and $\bise^{W, b, -\infty}: x \in [0, 1]^{\Omega_I} \mapsto (\conv{x}{W} < b) \in \Omega_I$.

\begin{theorem}[Dilation / Erosion Equivalence] \label{thm:check-dila}
    Let $W \in \R^{\Omega_S}$ be a set of weights, $b \in \R$ a bias and $p \in \R$ a scaling factor. Let $S \subset \Omega_S$ be a candidate SE. Given an almost binary input in $\almostbinary{u}{v}$\\
    
\begin{itemize}
    \item $\epsilon_{W, b, +\infty}$ is a dilation by $S$ if and only if $\epsilon_{W, b, -\infty}$ is an anti-dilation by $S$ if and only if 
    \begin{equation}\label{equation:dil-check}
        \sum_{i \in \compl{S} ~,~ w_i \geq 0}{w_i} + u \sum_{i \in S ~,~ w_i \geq 0}{w_i} \leq b < v \min_{i \in S}{w_i} + \sum_{i \in \Omega ~,~ w_i \leq 0}{w_i}
    \end{equation}
    \item $\epsilon_{W, b, +\infty}$ is an erosion by $S$ if and only if $\epsilon_{W, b, -\infty}$ is an anti-erosion by $S$ if and only if
    \begin{equation}\label{equation:ero-check}
        \sum_{i \in \Omega ~,~ w_i \geq 0}{w_i} - (1 - u)\min_{i \in S}{w_i}  \leq b < v\sum_{i \in S ~,~ w_i \geq 0}{w_i} + \sum_{i \in \Omega ~,~ w_i \leq 0}{w_i}
    \end{equation}
\end{itemize}

If one of these expressions is fulfilled, we say that the BiSE neuron is \textbf{activated}. Then we have $\forall i \in S ~,~ w_i \geq 0$ and $b \geq 0$, and the output of the BiSE $\bise_{W, b, p}$ is almost binary.
\end{theorem}


Equations \ref{equation:dil-check} and \ref{equation:ero-check} check if a candidate $S \subset \Omega_S$ activates the BiSE or not. If the BiSE is activated, given an almost binary input, the output is almost binary. Therefore, applying a BiSE on an almost binary image is the same as applying the morphological operator to the binary image. Finally, we can replace the entire BiSE neuron with the associated morphological operation in the activation case. The sign of $p$ determines if complementation is applied or not.

Theorem \ref{thm:check-dila} justifies a posteriori our enforcing of the weights and bias to be positive in the BiSE definition \ref{def:bise}: the goal is to achieve activation, which is only possible for positive parameters.

The BiSE can only be activated by at most one $S \subset \Omega_S$ simultaneously. We can recover this $S$ with proposition $\ref{prop:linear}$.

\begin{proposition}[Linear Check]\label{prop:linear}

Let us assume the BiSE is activated for almost binary images in $\almostbinary{u}{v}$.  Let $b \in \R$ be the BiSE bias, let $W \in \R^{\Omega_S}$ be the weights. 
Then there exists $\tau \in \R$ such that $S = \{i \in \Omega ~|~  W(i) \geq \tau\}$.

\begin{itemize}
    \item If the BiSE is a dilation / anti-dilation
\begin{equation}
\tau = \frac{1}{v}\big(b-\sum_{i \in \compl{S} ~,~ w_i \leq 0}{w_i}\big)
\end{equation}
    \item If the BiSE is an erosion / anti-erosion
\begin{equation}
\tau = \frac{1}{1 - u}\big(\sum_{i \in \Omega ~,~ w_i \geq 0}{w_i} - b\big)    
\end{equation}
\end{itemize}
\end{proposition}

We can check the two SEs associated with each threshold to recover the activation operation. If none works, then by contraposition, the BiSE is not activated. This is done in $\mathcal{O}(\card{\Omega})$ operations. 

\subsubsection{Binarization} \label{subsubsec:bise-binarization}

\paragraph{Formalism} The purpose of binarization is to replace the real parameters $(W, b, p)$ with binary parameters to improve inference efficiency \cite{simons2019review}. If the activation inequalities of Theorem $\ref{thm:check-dila}$ are respected, we saw that we could replace the BiSE with the morphological operation. That is a form of binarization: floating point numbers are no longer required. However, these inequalities are not necessarily respected. How to binarize the BiSE neuron in this case? We want to find the closest morphological operation. Let $(W^*, b^*, p^*) \in \R^{\Omega_S} \times \R_+ \times \R$ be the learned parameters. Let $A: (S, op) \in \partsof{\Omega_S} \times \{\text{dilation}, \text{erosion}\} \mapsto A(S, op) = \{(W, b) \in \R^{\card{\Omega_S} + 1} ~|~ f_S^{op}(W) \leq b < g_S^{op}(W)\}$, with $f_S^{op}$ and $g_S^{op}$ the bounds defined in Theorem $\ref{thm:check-dila}$. We want to define a dissimilarity function $d: (S, op), (W, b) \in (\partsof{\Omega_S} \times \{\text{dilation}, \text{erosion}\}) \times (\R^{\card{\Omega_S}} \times \R) \mapsto d(A(S, op), (W, b))$, and we want to find
\begin{equation}
    \argmin{(S, op) \in \partsof{\Omega_S} \times \{\text{dilation}, \text{erosion}\}}{d(A(S, op), (W^*, b^*))}
\end{equation}

\paragraph{Choice of dissimilarity} A first choice is to use:
\begin{equation}
    d(A(S, op), (W, b)) = \max\Big(0, f_S^{op}(W) - b, b - g_S^{op}(W)\Big)
\end{equation}
The search for $(S, op) \in \partsof{\Omega_S} \times \{\text{dilation}, \text{erosion}\}$ has exponential complexity $\mathcal{O}(2^{\card{\Omega_S}}$). To simplify this search, we take inspiration from proposition \ref{prop:linear}: if the BiSE is activated, we know that $S = \{i ~|~ w_i \geq \tau\}$ for a certain $\tau$. Therefore, we reduce the search to all $S$ of this thresholded form, with $\tau \in \{w_i ~|~ i \in \Omega_S\}$. This reduces the complexity to $\mathcal{O}(\card{\Omega_S})$.

\subsection{Binary Structuring Element Layer}

One strength of ConvNets is their ability to learn multiple filters per layer, which we also want to ensure. In a convolutional layer, each final channel is a sum of one filter per input channel. In our case, the final result is either a union or an intersection of the elementary operators (dilation, erosion, anti-dilation, anti-erosion). Therefore, we want a layer that can learn the union or intersection of any combination of inputs. Let us consider $n$ binary images $x_1, ..., x_n \subset \Omega$. Let $\cC \subset \discint{1}{n}$. Then the intersection and union are given as: 

\begin{align}
\indicator{\bigcap_{i \in \cC}{x_i}} &= \Big( \sum_{i \in \cC}{\indicator{x_i}} \geq \card{\cC}\Big)\\
\indicator{\bigcup_{i \in \cC}{x_i}} &= \Big( \sum_{i \in \cC}{\indicator{x_i}} \geq 1\Big)
\end{align}

As with the BiSE, we can use a single scalar to discriminate between the union or the intersection. To learn the set $\cC$, we can use a parameter $\beta_i$ for each image. The following definition ensues:

\begin{definition}[$\lui$]
 Let $\beta = (\beta_1, ..., \beta_c) \in \R^c$.  Let $b \in \R$ be a bias and $p \in \R$ a scaling factor. We define the \textit{$\lui$} (Layer Intersection Union) as a thresholded linear combination:
 
 \begin{equation}
   \lui^{\beta, b, p}: x \in (\Z^d)^c \mapsto \xi\bigg(p \Big(\sum_{i = 1}^c{f^+(\beta_i)x_i} - f^+(b) \Big)\bigg) \in \Z^d
 \end{equation}
\end{definition}

A $\lui$ layer can learn any intersection or union of any number of almost binary inputs. It is a particular case of BiSE layer. In this case, the SE support would be $\Omega_{lui} = \{0\}^d \times \discint{1}{n}$. Learning the intersection is equivalent to learning an erosion, and learning the union is equivalent to learning a dilation. Then, we can deduce the following Theorem, by using the BiSE Theorem \ref{thm:check-dila}. We denote by $\I = \bigtimes_{k=1}^n{\almostbinary{u_k}{v_k}}$ the set of images with $n$ almost binary channels.

\begin{theorem}[$\lui$ intersection / union equivalence]\label{thm:check-lui}

Let $n \in \N^*$ and $\cC \subset \discint{1}{n}$. Let $b \in \R$. Let $(u_i < v_i) \in [0,1]^{2n}$. Let $\beta \in \R^n$.

\begin{itemize}
    \item $\lui^{\beta, b, +\infty}$ is an intersection by $\cC$ if and only if
    \begin{equation}
        \sum_{k=1~,~ \beta_k \geq 0}^n{\beta_k} - \min_{k\in \cC}{\Big[(1-u_k)\beta_k\Big]} \leq b < \sum_{k\in \cC ~,~ \beta_k \geq 0}{\beta_k v_k} + \sum_{k = 1 ~,~ \beta_k \leq 0}^n{\beta_k}
    \end{equation}
    
    \item $\lui^{\beta, b, +\infty}$ is a union by $\cC$ if and only if 
    \begin{equation}
        \sum_{k \in \cC ~,~ \beta_k \geq 0}{\beta_k u_k} + \sum_{k \in \discint{1}{n} ~\backslash~ \cC ~,~ \beta_k \geq 0}{\beta_k} \leq b < \min_{k \in \cC}(\beta_kv_k) + \sum_{k = 1 ~,~ \beta_k \leq 0}^n{\beta_k}
    \end{equation}
    
\end{itemize}

If one of these inequalities is respected, we say that the $\lui$ is \textbf{activated}. Then, we have $\forall k \in \cC ~,~ \beta_k \geq 0$ and $b \geq 0$ and the output of the $\lui$ is almost binary.

\end{theorem}

As for the BiSE, if the $\lui$ is activated, the set $\cC$ can be found by thresholding the $(\beta_k)_{k}$ for a specific value. If the $\lui$ is not activated, it is still possible to binarize with the same method described for the BiSE in section \ref{subsubsec:bise-binarization}.


We combine the BiSE neurons and the $\lui$ in order to be able to learn the morphological operators and aggregate them as unions or intersections.

\begin{definition}[BiSEL]
A BiSEL (BiSE Layer) is the combination of multiple BiSE and multiple $\lui$.
Let $(\bise_{n, k})$ be $N*K$ BiSE and $(\lui_k)_k$ be $K$ $\lui$. Then we define a BiSEL as:

\begin{equation}
\bisel : x \in (\Z^d)^n \mapsto \bigg(\lui_k\Big[\big(\bise_{n, k}(x_n)\big)_n\Big]\bigg)_k \in (\Z^d)^K
\end{equation}

The LUI and BiSE are called the BiSEL's \textbf{elements}. If every element is activated, we say that the BiSEL is \textbf{totally activated}.

\end{definition}

If the BiSEL is activated,  given almost binary inputs, the outputs of the BiSEL are also almost binary. 

\begin{figure}[h]
    \centering
    \includegraphics[width=.4\linewidth]{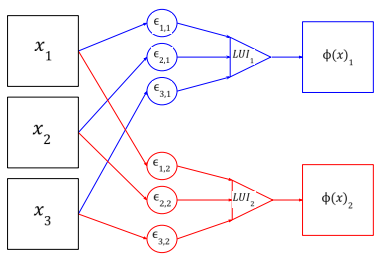}
    \caption{Schema of BiSEL. Input $x$ with 3 channels. Output $\phi(x)$ with 2 channels.}
    \label{fig:bisel}
\end{figure}

The BiSEL follows the same logic as a convolutional layer. In ConvNets, at each layer, we have $k$ filters. Each filter applies one convolution to each channel, and then we apply a linear combination to all convoluted channels. In BiSEL, we have $K$ filters (the number of $\lui$). For each filter, we apply a morphological operation to a channel, and then we aggregate by taking the intersection or union of any channels. A schema of BiSEL can be found in figure \ref{fig:bisel}.

Binarizing a BiSEL is equivalent to binarizing each of its element. If an element is activated, it is easy to binarize it. For those who are not, we must approximate them.

The parameters are the BiSE weights $\{W_{n, k}\}_{n, k} \subset \R^{\Omega_S}$, the BiSE biases $\{b_{n, k}\}_{n, k} \subset \R$, the $\lui$ parameters $\{\beta_{n, k}\}_{n, k} \subset \R$ and the $\lui$ biases $\{b^l_k\}_k \subset \R$. There are $N\cdot K(\card{\Omega_S} + 2) + K$ parameters.

\subsection{BiMoNN}


The BiSEL is akin to a complete convolution layer with multiple filters. We can simply stack multiple BiSELs to create a network similar to a convolutional neural network.

\begin{definition}[BiMoNN]
Let $\bisel_1, ..., \bisel_L$ be $L$ BiSEL. We define the Binary Morphological Neural Network ($\bimonn$) as:
\begin{equation}
    \bimonn = \bisel_L \circ ... \circ \bisel_1
\end{equation}
We say that the $\bimonn$ is \textbf{totally activated} if all the composing BiSELs are totally activated.
\end{definition}

The BiMoNN can learn the composition of any intersection or union of dilation, erosion, anti-dilation, and anti-erosion. That includes but is not limited to opening, closing, black top-hat, and white top-hat.

There are two drawbacks to our current formulation. First, the input has the same dimension as the output. ConvNets can be combined in multi-scale architectures, which compresses the data. Currently, we cannot solve classification tasks with our method. Second, we can only learn predeternined composition of operators. However, some morphological operators like skeletonization or reconstruction need an undetermined number of iterations before convergence. Incorporating these features is left for further work.


\paragraph{Learning phase} The $\bimonn$ is fully differentiable, and we learn its parameters with classical deep learning techniques. Let $\mathcal{L}: \Omega_I^2 \rightarrow \R$ be a differentiable loss function. Depending on our interpretation of the outside domain $\Z^d ~\backslash~ \Omega_I$, the effects on the border can vary. Thus, the loss is computed by avoiding the borders, whose size is half of the kernel size $\Omega_S$ in each direction. We try classical segmentation losses: the dice loss \cite{milletari2016vnet}, the binary cross-entropy (BCE), and the mean squared error (MSE).

Then, given a dataset of $N$ labeled images $(X, Y)$,  we minimize the error $\frac{1}{N}\sum_{i=1}^N{\mathcal{L}(\bimonn(x_i), y_i)}$ using Adam~\cite{kingma2017adam}. We compute the loss gradient with the backpropagation algorithm \cite{Rumelhart:1986we}. In practice, we replace $f^+(b)$ by $f^+(b) + 0.5$ to help with training for all BiSE and $\lui$ operators. The convolution weights $W$ follow the kaiming uniform initialization~\cite{glorot2010understanding}, and the biases begin at $f^+(2) + 0.5 = 0.63$. We initialize The BiSE and $\lui$ scaling factors $p$ at $0$. That may not be optimal, as it could bias the BiSE towards some operations and Sels. Could we find other unbiased methods toward a particular operation or Sel ? This is left for further research.

\paragraph{Binarization} Once the training is done, we can binarize each element independently to binarize the whole $\bimonn$.
Theorems \ref{thm:check-dila} and \ref{thm:check-lui} give activation conditions regarding a specific set of almost binary images $\almostbinary{u}{v}$. Given a layer $l$, the range $[u_l, v_l]$ depends on the activation of the layer $l-1$ and only exists if all the layers $\bisel_1, ..., \bisel_{l-1}$ are totally activated. This ideal situation does not occur in most cases. 
In practice, we binarize the network sequentially layer by layer, from the first to the last, and compute the almost binary range $[u_l, v_l]$ along the way. If a layer $\phi_l$ is not totally activated, we use the approximation described in \ref{subsubsec:bise-binarization}: then, the next range becomes $[u_{l+1}, v_{l+1}] = [0, 1]$.

\section{Experiments}
\label{sec:experiments}

\subsection{Classical morphological operators}

We will be using the three datasets described in \cite{aouad2022binary}: \diskorect{}, a set of custom-generated images composed of ellipses and rectangles that create as many different situations as possible; thresholded \mnist{} upsized to $50 \times 50$, and its complementation that we call inverted \mnist{}. Examples are shown in figure \ref{fig:dataset_example}.



\begin{figure}
    \centering
        \includegraphics[width=.07\linewidth]{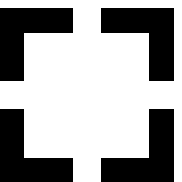}
        \includegraphics[width=.07\linewidth]{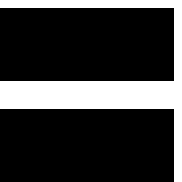}
        \includegraphics[width=.07\linewidth]{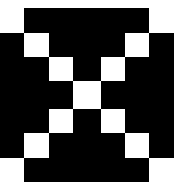}
    \caption{Target SEs}
    \label{fig:target_sels}
\end{figure}

\paragraph{Experiment description}We learn the following operators: dilation, erosion, opening, closing, black top-hat, and white top-hat. The 3 SEs used can be seen on table \ref{fig:target_sels}. For erosion and dilation on \mnist{}, we use smaller SEs of size $5 \times 5$. Else for all other operations, SEs are of size $7 \times 7$. 
The experimental protocol is as follows: let $\gamma$ be one of these morphological operators. The input-output couples are $\{(x_i, \gamma(x_i)) ~|~ x_i \in \text{dataset}\}$. The hyperparameters are the loss used, the learning rate, and the batch size. For each result, see the code for the hyperparameters used. We measure the DICE score \cite{dice1945measures} between the predictions and targets.

\subsection{A real example: defining regions of interest}

Axial Spondyloarthritis (axSpA) is the most common auto-immune rheumatic disease. It is painful and debilitating if untreated or incorrectly diagnosed. The disease can be detected using MRIs: the inflammation is located on the sacroiliac joint. Given the sacrum and the iliac shapes, the task is to detect this joint region. We call this task axSpA.

\begin{figure}[h]
    \centering
    \begin{subfigure}{.32\linewidth}
        \includegraphics[width=.48\textwidth]{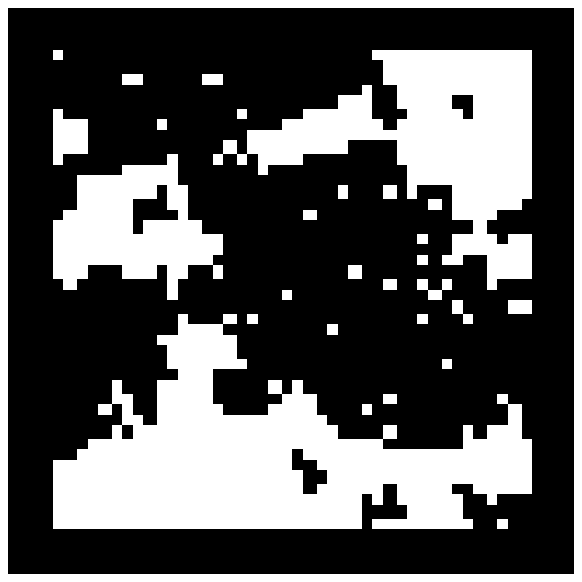} \includegraphics[width=.48\textwidth]{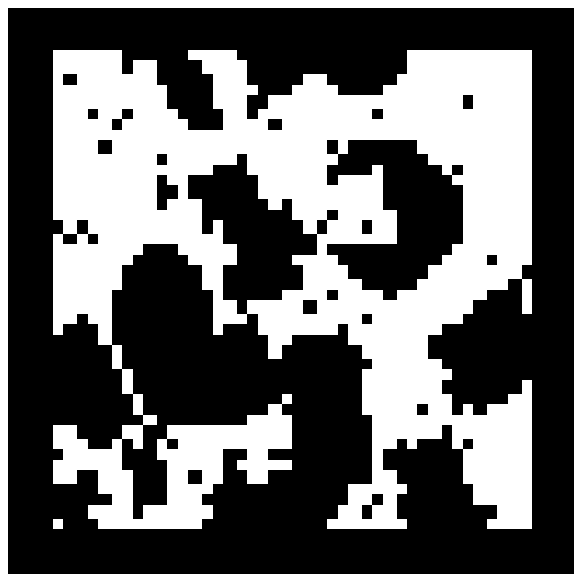}
        \caption{Diskorect}
        \label{subfig:diskorect}
    \end{subfigure}
    \begin{subfigure}{.165\linewidth}
        \includegraphics[width=\linewidth]{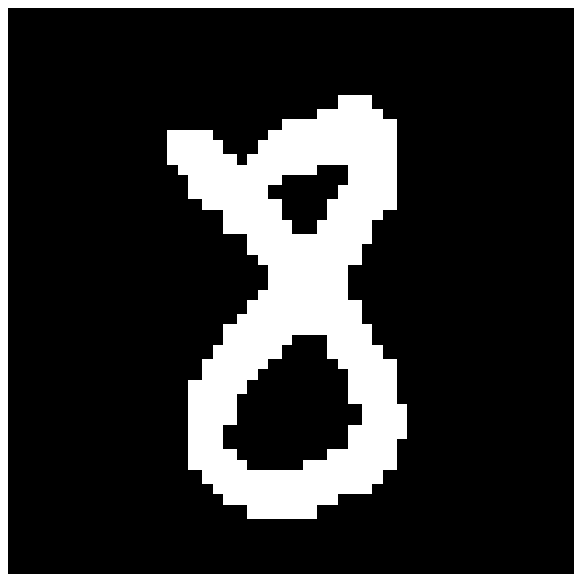}
        \caption{MNIST}
        \label{subfig:mnist}
    \end{subfigure}
    \begin{subfigure}{.165\linewidth}
        \includegraphics[width=\linewidth]{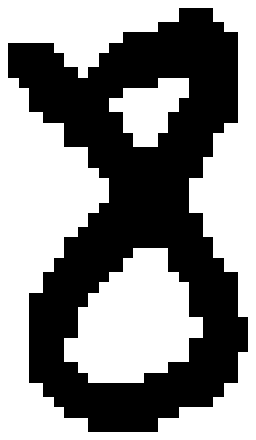}
        \caption{Inverted MNIST}
        \label{subfig:inverted_mnist}
    \end{subfigure}
    \begin{subfigure}{.32\linewidth}
        \includegraphics[width=.48\textwidth]{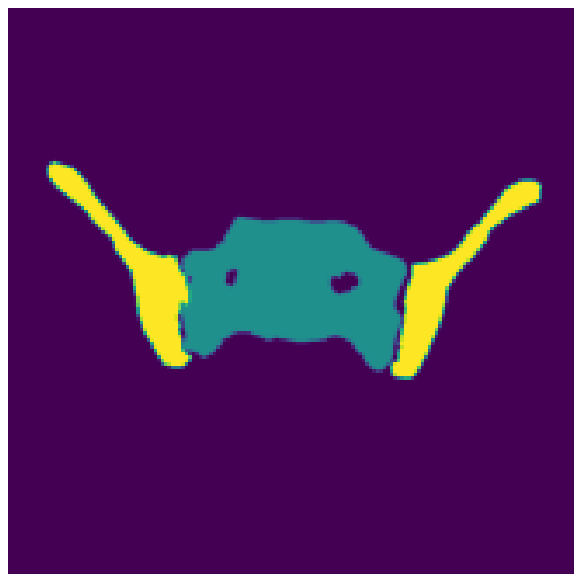}
        \includegraphics[width=.48\textwidth]{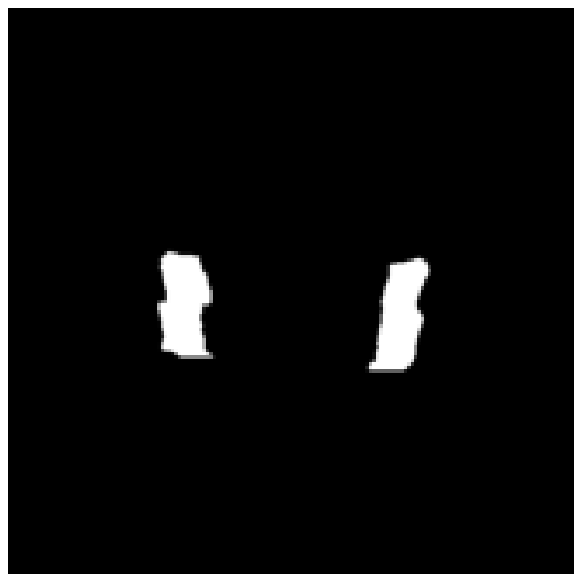}
        \caption{axSpA input (left) and target (right)}
        \label{subfig:axspa-ex}
    \end{subfigure}
    \caption{Datasets example}
    \label{fig:dataset_example}
\end{figure}

The dataset is composed of around 1100 images with categorical images of size about $512 \times 512$ with two values: one for the iliac and one for the sacrum. We transform them into binary images by separating each categorical value into one channel: final images have 2 channels. The target is defined as the joint region. Let $S_a = \discint{-20}{20} \times \{0\}$. The target operation is $\gamma_a: X \in \{0, 1\}^{512 \times 512 \times 2} = (\dil{X_1}{S_a}) \cap (\dil{X_2}{S_a})$, with $X_i$ the $i-$channel. The resulting input-output couples are $\{(x_i, \gamma_a(x_i)) ~|~ x_i \in \text{axSpA dataset}\}$ (see figure \ref{subfig:axspa-ex}).

\section{Discussion}
\label{sec:discussion}

Results on all operations are depicted in table \ref{tab:results-parallel}.

\begin{table}[h]
    \centering
    \caption{DICE Results on morphological operations. $\R$ means real weights, $\mathbb{B}$ means after binarization (or approximate). \textbf{Bold} means DICE $> 0.8$.}
    \resizebox{\linewidth}{!}{\newcommand{\tophatelt}[2]{
    \makecell{$\R ~ #1$ \\ $\mathbb{B} ~ #2$}
}        

    \begin{tabular}{| c | c | c | c | c | c | c | c | c | c |}
    \hline
         Dataset & \multicolumn{3}{c|}{Diskorect}  & \multicolumn{3}{c|}{MNIST} & \multicolumn{3}{c|}{Inverted MNIST} \\
        \hline
        Operation $\backslash$ Sel & Disk & Stick & Cross & Disk & Stick & Cross & Disk & Stick & Cross \\
        \hline
        Dilation $\dil{}{}$ & \tophatelt{\mathbf{1.00}}{\mathbf{1.00}}  & \tophatelt{\mathbf{1.00}}{\mathbf{1.00}}  & \tophatelt{\mathbf{1.00}}{\mathbf{1.00}}  & \tophatelt{\mathbf{1.00}}{\mathbf{1.00}}  & \tophatelt{\mathbf{1.00}}{\mathbf{1.00}}  & \tophatelt{\mathbf{1.00}}{\mathbf{1.00}}  & \tophatelt{\mathbf{1.00}}{\mathbf{1.00}}  & \tophatelt{\mathbf{1.00}}{\mathbf{1.00}}  & \tophatelt{\mathbf{1.00}}{\mathbf{1.00}} \\
        \hline
        Erosion $\ero{}{}$ & \tophatelt{\mathbf{1.00}}{\mathbf{1.00}}  & \tophatelt{\mathbf{1.00}}{\mathbf{1.00}}  & \tophatelt{\mathbf{1.00}}{\mathbf{1.00}}  & \tophatelt{\mathbf{1.00}}{\mathbf{1.00}}  & \tophatelt{\mathbf{1.00}}{\mathbf{1.00}}  & \tophatelt{\mathbf{1.00}}{\mathbf{1.00}}  & \tophatelt{\mathbf{1.00}}{\mathbf{1.00}}  & \tophatelt{\mathbf{1.00}}{\mathbf{1.00}}  & \tophatelt{\mathbf{1.00}}{\mathbf{1.00}} \\
        \hline
        Opening $\whitehat{}{}$ & \tophatelt{\mathbf{1.00}}{\mathbf{1.00}}  & \tophatelt{\mathbf{1.00}}{\mathbf{1.00}}  & \tophatelt{\mathbf{1.00}}{\mathbf{1.00}}  & \tophatelt{\mathbf{1.00}}{\mathbf{1.00}}  & \tophatelt{\mathbf{1.00}}{\mathbf{0.99}}  & \tophatelt{\mathbf{1.00}}{\mathbf{0.99}}  & \tophatelt{\mathbf{0.99}}{\mathbf{0.93}}  & \tophatelt{\mathbf{1.00}}{\mathbf{0.98}}  & \tophatelt{\mathbf{0.99}}{\mathbf{0.90}} \\
        \hline
        Closing $\whitehat{}{}$ & \tophatelt{\mathbf{1.00}}{\mathbf{1.00}}  & \tophatelt{\mathbf{0.92}}{\mathbf{0.87}}  & \tophatelt{\mathbf{1.00}}{\mathbf{1.00}}  & \tophatelt{\mathbf{1.00}}{\mathbf{1.00}}  & \tophatelt{\mathbf{1.00}}{0.66}  & \tophatelt{\mathbf{1.00}}{\mathbf{1.00}}  & \tophatelt{\mathbf{1.00}}{\mathbf{1.00}}  & \tophatelt{\mathbf{1.00}}{\mathbf{0.91}}  & \tophatelt{\mathbf{1.00}}{\mathbf{1.00}} \\
        \hline
        White tophat $\whitehat{}{}$ & \tophatelt{\mathbf{0.82}}{0.28}  & \tophatelt{\mathbf{0.97}}{0.33}  & \tophatelt{\mathbf{0.81}}{0.80}  & \tophatelt{0.00}{0.40}  & \tophatelt{0.06}{0.32}  & \tophatelt{0.78}{0.65}  & \tophatelt{0.00}{0.12}  & \tophatelt{0.13}{0.14}  & \tophatelt{0.22}{0.21} \\
        \hline
        Black tophat $\blackhat{}{}$ & \tophatelt{\mathbf{1.00}}{0.31}  & \tophatelt{\mathbf{1.00}}{\mathbf{0.82}}  & \tophatelt{\mathbf{1.00}}{\mathbf{1.00}}  & \tophatelt{0.00}{0.00}  & \tophatelt{0.38}{0.00}  & \tophatelt{0.00}{0.00}  & \tophatelt{\mathbf{0.98}}{0.76}  & \tophatelt{\mathbf{1.00}}{0.29}  & \tophatelt{\mathbf{1.00}}{0.26} \\
        \hline
    \end{tabular}

}
    \label{tab:results-parallel}
\end{table}

\subsection{Sequential operations}

For dilation and erosion, the SEs are perfectly learned. The perfect DICE (=1) is reached in almost all cases. Even when the BiSEs are not activated, the approximate binarizations yield good results. Performance can decrease (see disk erosion on inverted \mnist{}), or increase (see disk erosion on \mnist{}).

On opening and closing, we reach perfect dice for most cases. All SEs can be seen on the code repository. We manage to recover the target SEs for most operations except on the opening on the inverted \mnist{}. We want to test the following hypothesis $\mathcal{H}$: dual operators learn similarly on dual datasets. On \diskorect{}, for any image $X$ in this dataset, then $\compl{X}$ is also in it. Therefore $\mathcal{H}$ implies that opening and closing should have similar results. Already the hypothesis is rejected if we look at the stick opening vs. the stick closing: one succeeds and the other fails. On \mnist{}, $\mathcal{H}$ implies that the closing on \mnist{} should behave the same as the opening on inverted \mnist{} and vice versa. That is not observed and refutes further the hypothesis. Note that the model learns two anti-dilations for the closing, which is equivalent to learning dilation followed by erosion.

We compare our results to $\mathcal{L}Morph$ and $\mathcal{S}Morph$ \cite{kirszenberg2021going}. They reach almost perfect DICE for erosion and dilation, but recovering the SEs from the learned weights is not straightforward. Moreover, on the opening and closing, the $\mathcal{L}Morph$ presents stability issues, and the $\mathcal{S}Morph$ fails to converge correctly to the proper operations. Those results are not surprising as these networks were designed for grey-scale morphology.

\subsection{Parallel operations}

\begin{figure}
    \centering
    \begin{subfigure}{.48\linewidth}
        \includegraphics[width=\linewidth]{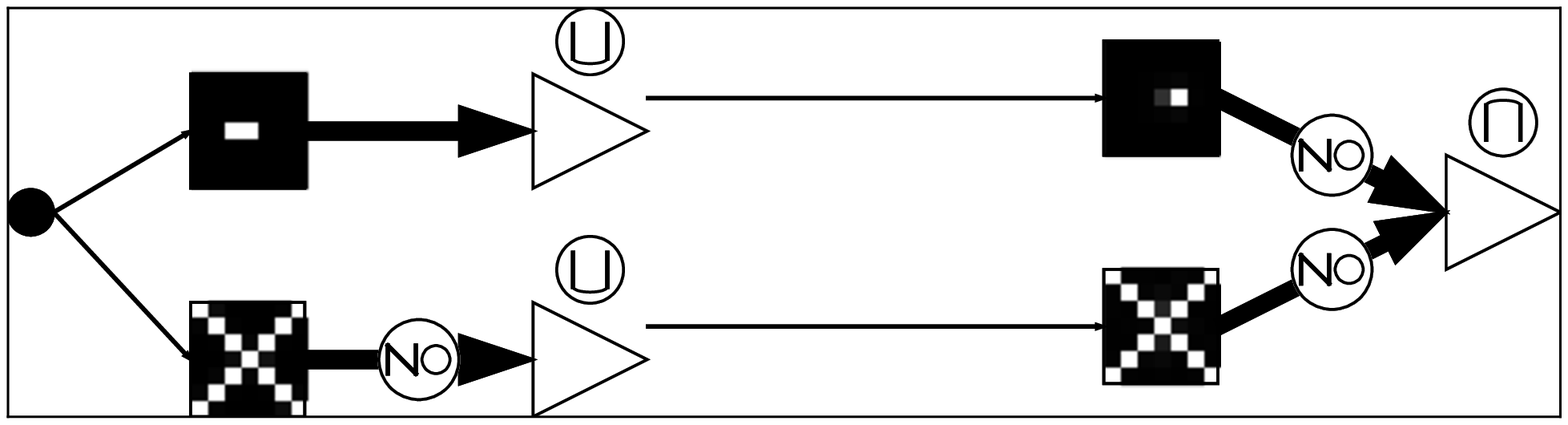}
        \caption{Weights}
    \end{subfigure}
    \begin{subfigure}{.48\linewidth}
        \includegraphics[width=\linewidth]{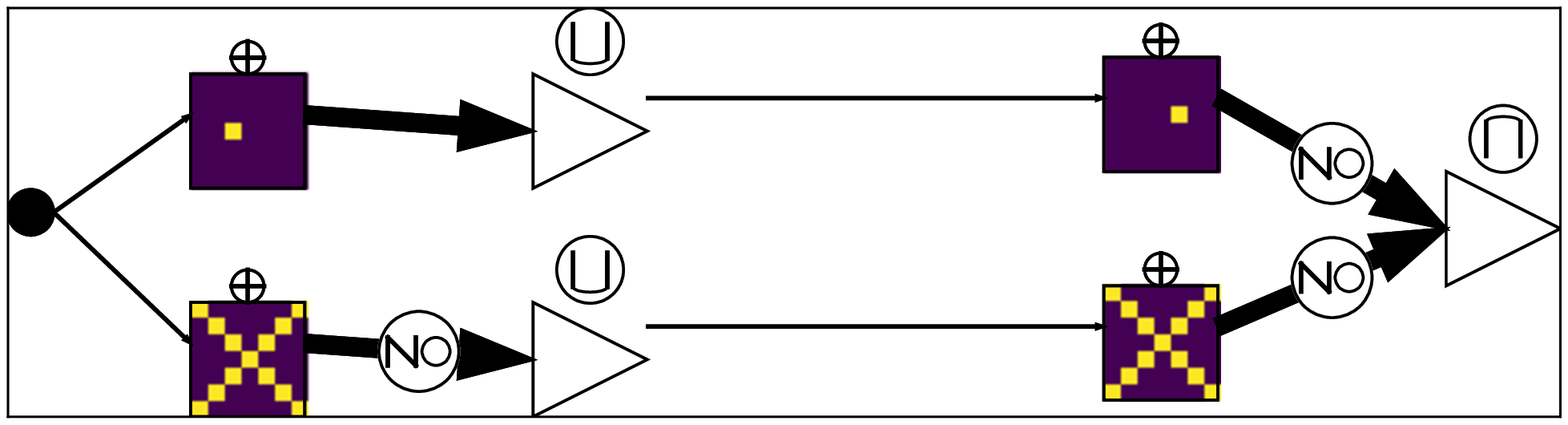}
        \caption{Binarization}
    \end{subfigure}
    \caption{Learned model for black top-hat with cross}
    \label{fig:black_top_hat_cross}
\end{figure}

In this section, we study the model's ability to learn the complementation and intersection with top-hat transform.
We manage to learn the black top-hat for the \diskorect{} and inverted \mnist{}. For the cross, the binarization is perfect (see figure \ref{fig:black_top_hat_cross}). The upper branch learns the complementation of the identity, and the bottom branch learns the closing with two sequential anti-dilations. The white top-hat only worked on \diskorect{}. The binarization decreases performance.

\begin{figure}
    \centering
    \begin{subfigure}{.3\linewidth}
        \includegraphics[width=\linewidth]{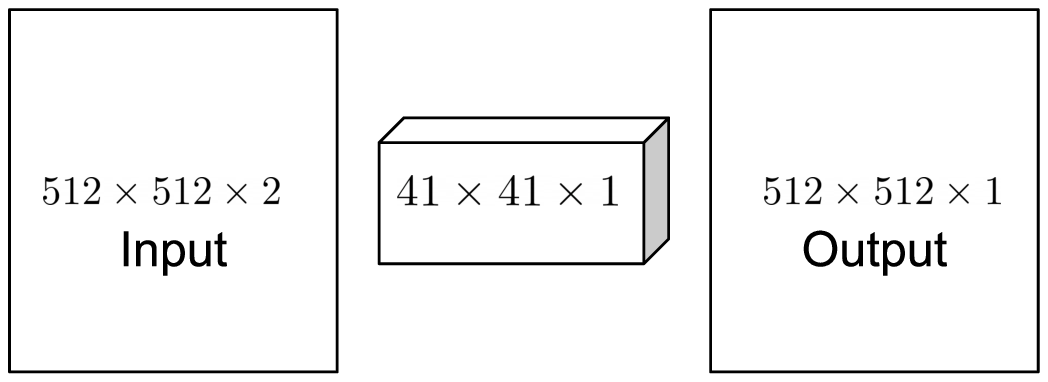}
        \caption{Architecture 1}
    \end{subfigure}
    \hspace{.15\linewidth}
    \begin{subfigure}{.4\linewidth}
        \includegraphics[width=\linewidth]{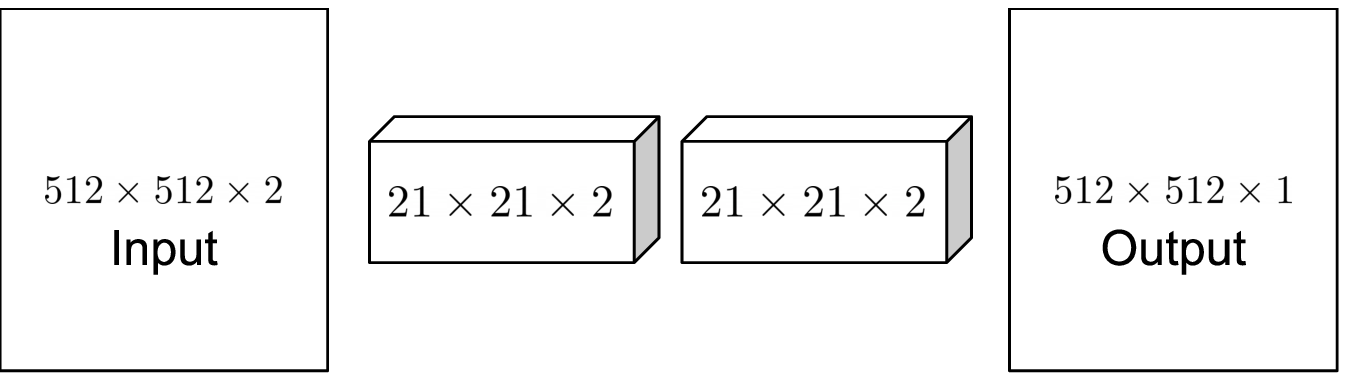}
        \caption{Architecture 2}
    \end{subfigure}
    \begin{subfigure}{.6\linewidth}
        \includegraphics[width=\linewidth]{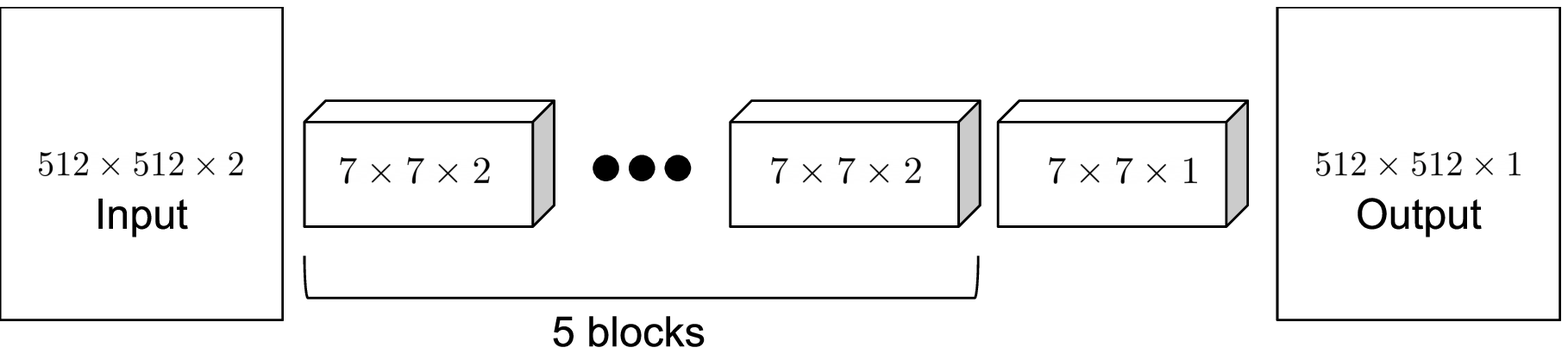}
        \caption{Architecture 3}
    \end{subfigure}
    \caption{Architectures used for axSpA}
    \label{fig:axspa-architectures}
\end{figure}

\subsection{AxSpA regions of interest}

We attempt to learn the joint region. We want to learn the intersection of two dilations of size $41 \times 41$. We try three different architectures 
that can theoretically learn this operation (see figure \ref{fig:axspa-architectures}). They go from wide and shallow to deeper and narrower. Architectures 1 and 2 learn the operation well, reaching an excellent DICE ($1$ and $0.9$). 
Architecture 1's weights are close to the target, and the binarization yields good results (DICE=$0.98$). On the other hand, architecture 2 learns wider dilations, and its binarization is quite bad (DICE=$0.7$). Note that there are multiple ways for this architecture to learn the target. Finally, the deeper network failed to learn anything. That may be due to the initialization: the output is constant and stuck in a zero-grad zone, which is a case of vanishing gradient. This application is hopeful in the study of axSpA. This model could be used with classical convolutional neural networks to improve MRI analysis by exploiting the ROI's detection using segmentation results.

\section{Conclusion}
\label{sec:print}

We create a neural network designed to work on binary elements. It comes with explainable results that can be used to binarize the network, improving its embedded applicability for inference. We manage to learn the erosion, dilation, opening, and closing of small structuring elements. We can learn sequential operations and more complex ones that need the intersection of multiple operators. Even with bigger structuring elements, we learn appropriate behaviors for concrete applications in medical imaging. However, there are still unanswered questions. Why does the training differ for dual operations? How can we avoid vanishing gradient in deeper networks? How can we reduce the dimension to create a classification model? Can we redesign the model to take into account undetermined number of iterations? How can we binarize more efficiently to preserve the performance, especially if there are multiple channels?

%
%
%

\bibliographystyle{splncs04}
\bibliography{mybibliography}

\appendix
\begin{table}[h]
    \centering
    \caption{Results on erosion and dilation. DICE error ($1 - \text{DICE}$) is presented for each case: $\R$ before binarization, $\mathbb{B}$ after binarization. \textcolor{blue}{\checkmark} indicates that the neuron is activated.}
    \resizebox{\linewidth}{!}{\newcommand{\dileroelt}[4]{
    \begin{minipage}{.12\linewidth}
        \ifstrequal{#2}{ok}{
          \center{\textcolor{blue}{\checkmark}}
          }{
            \center{\textcolor{red}{$\times$}}
        }
          \includegraphics[width=\textwidth]{#1}
          $\R ~~ #3$
          $\mathbb{B} ~~ #4$
    \end{minipage}
}

     \begin{tabular}{ | c | c | c | c | c | c | c | c | c | c |}
        \hline
        Dataset & \multicolumn{3}{c|}{Diskorect}  & \multicolumn{3}{c|}{MNIST} & \multicolumn{3}{c|}{Inverted MNIST} \\
        \hline
        Operation & Disk & Stick & Cross & Disk & Stick & Cross  & Disk & Stick & Cross \\
        \hline
        Target
        & \begin{minipage}{.12\linewidth}
          \includegraphics[width=\textwidth,left]{selem_results/true_disk7.eps}
        \end{minipage}
        & \begin{minipage}{.12\linewidth}
          \includegraphics[width=\textwidth]{selem_results/true_hstick7.eps}
        \end{minipage}
        & \begin{minipage}{.12\linewidth}
          \includegraphics[width=\textwidth]{selem_results/true_dcross7.eps}
        \end{minipage}
        & \begin{minipage}{.12\linewidth}
          \includegraphics[width=\textwidth]{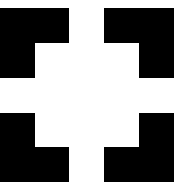}
        \end{minipage}
        & \begin{minipage}{.12\linewidth}
          \includegraphics[width=\textwidth]{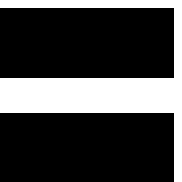}
        \end{minipage}
        & \begin{minipage}{.12\linewidth}
          \includegraphics[width=\textwidth]{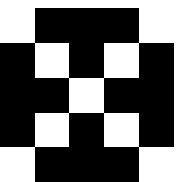}
        \end{minipage}
        & \begin{minipage}{.12\linewidth}
          \includegraphics[width=\textwidth]{selem_results/true_disk5.eps}
        \end{minipage}
        & \begin{minipage}{.12\linewidth}
          \includegraphics[width=\textwidth]{selem_results/true_hstick5.eps}
        \end{minipage}
        & \begin{minipage}{.12\linewidth}
          \includegraphics[width=\textwidth]{selem_results/true_dcross5.eps}
        \end{minipage}
        \\        
        \hline
        Dilation $\dil{}{}$
        & 
        \dileroelt{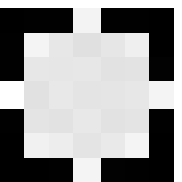}{ok}{0.000}{0.000}
        & 
        \dileroelt{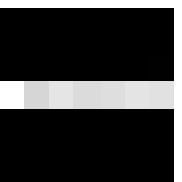}{ok}{0.000}{0.000}
        & 
        \dileroelt{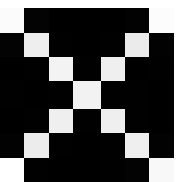}{ok}{0.000}{0.000}
        & 
        \dileroelt{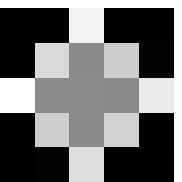}{ok}{0.000}{0.000}
        & 
        \dileroelt{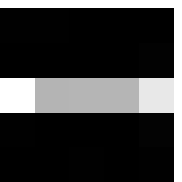}{ok}{0.000}{0.000}
        & 
        \dileroelt{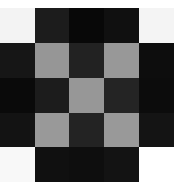}{ko}{0.000}{0.000}
        & 
        \dileroelt{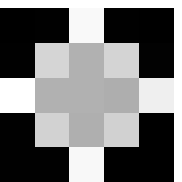}{ok}{0.000}{0.000}
        & 
        \dileroelt{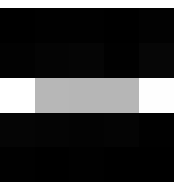}{ok}{0.000}{0.000}
        & 
        \dileroelt{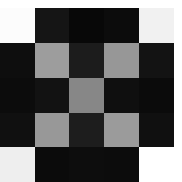}{ko}{0.000}{0.000}
                
        \\       
        \hline
        Erosion $\ero{}{}$
        & 
        \dileroelt{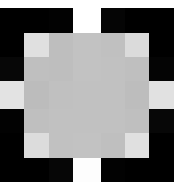}{ko}{0.000}{0.000}
        & 
        \dileroelt{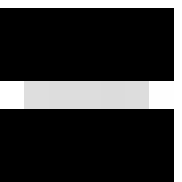}{ok}{0.000}{0.000}
        & 
        \dileroelt{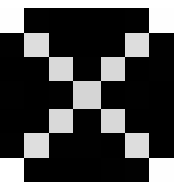}{ok}{0.000}{0.000}
        & 
        \dileroelt{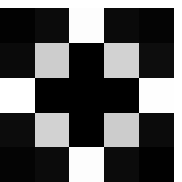}{ok}{0.001}{0.000}
        & 
        \dileroelt{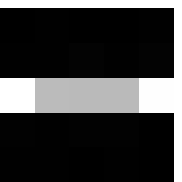}{ok}{0.000}{0.000}
        & 
        \dileroelt{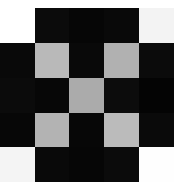}{ko}{0.000}{0.000}
        & 
        \dileroelt{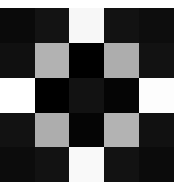}{ko}{0.000}{0.004}
        & 
        \dileroelt{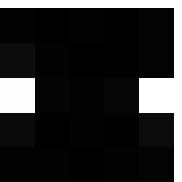}{ok}{0.001}{0.001}
        & 
        \dileroelt{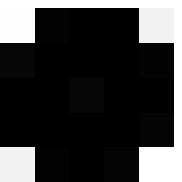}{ok}{0.000}{0.000}
        \\
        \hline
      \end{tabular}

}    
    \label{tab:erodila}
\end{table}

\begin{figure}[h]
    \centering
    \begin{subfigure}{\linewidth}
        \includegraphics[width=\linewidth]{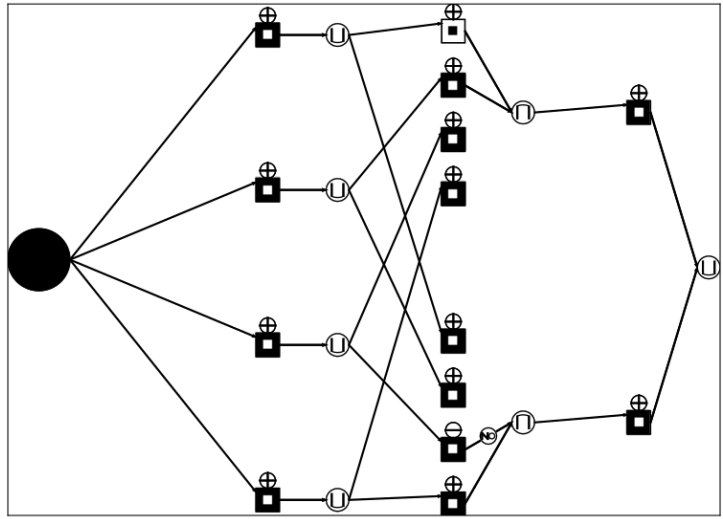}
        \caption{Elementary operators}
        \label{subfig:isolated-target}
    \end{subfigure}
    \begin{subfigure}{\linewidth}
        \includegraphics[width=\linewidth]{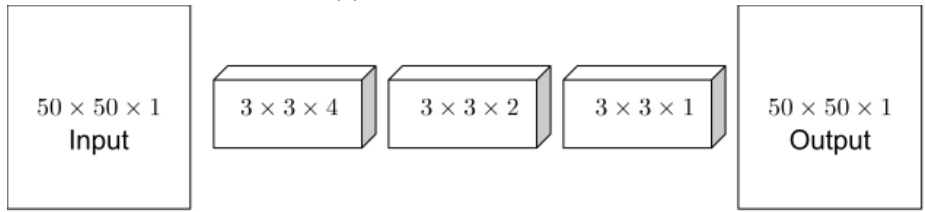}
        \caption{Corresponding architecture}
        \label{subfig:isolated-architecture}
    \end{subfigure}
    \caption{Bernoulli Denoising Set Up}
    \label{fig:bernoulli-denoising-target}
\end{figure}

\begin{figure}
    \centering
    \includegraphics[width=\linewidth]{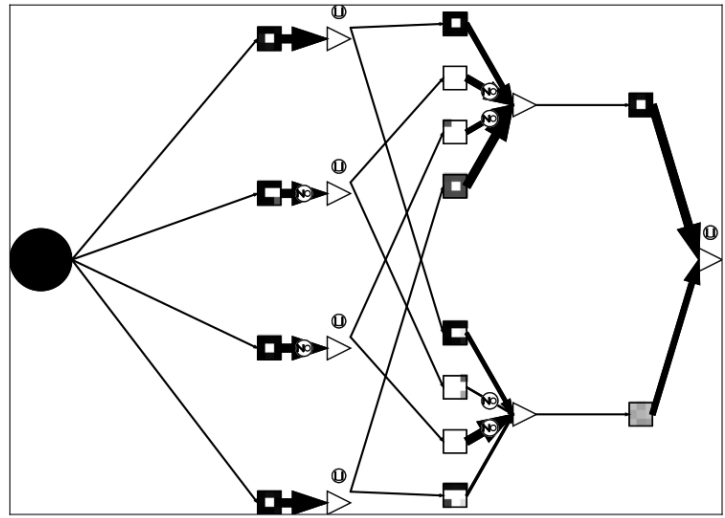}
    \caption{Bernoulli Denoising Results, weights learned}
    \label{subfig:isolated-weights}
\end{figure}

\begin{figure}
    \centering
    \includegraphics[width=\linewidth]{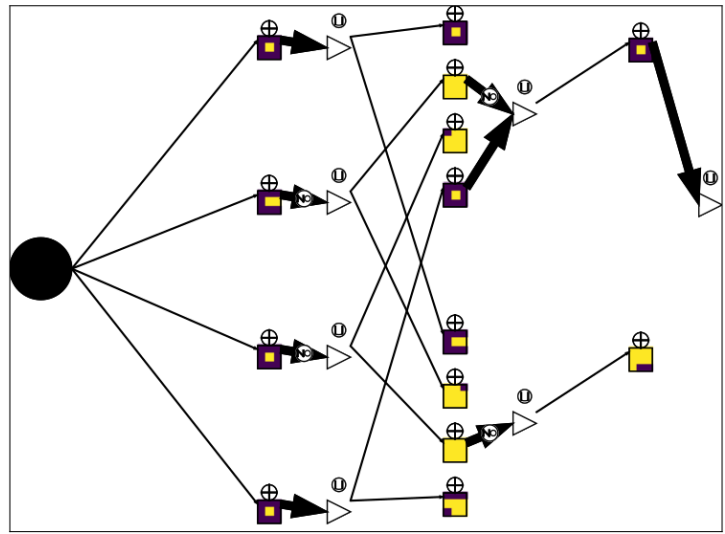}
    \caption{Bernoulli Denoising Results, binarization}
    \label{fig:bernoulli-denoising-results}
\end{figure}

\begin{figure}[h]
    \centering
    \begin{subfigure}{.45\linewidth}
        \includegraphics[width=\linewidth]{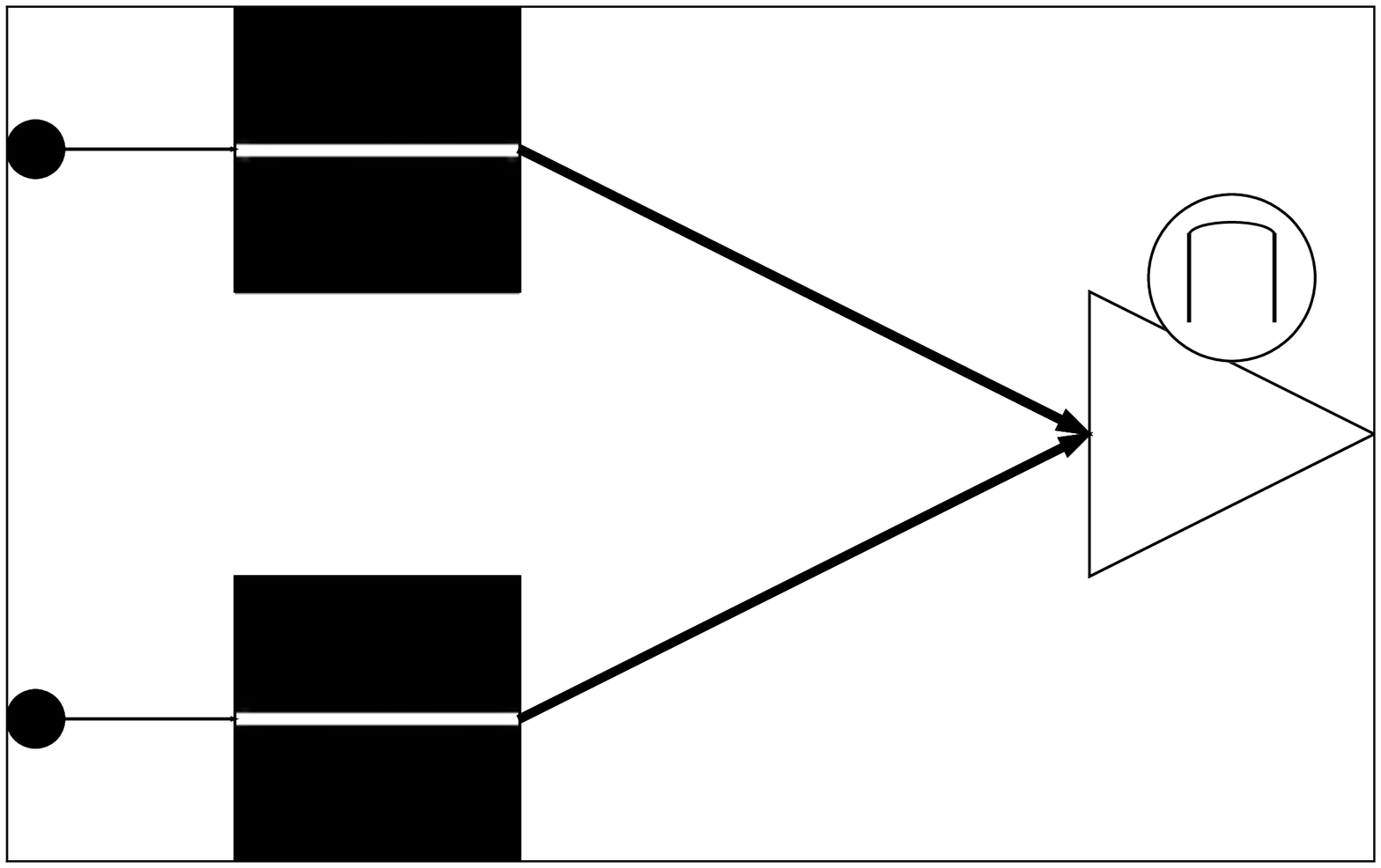}
        \caption{Weights for architecture 1}
    \end{subfigure}
    \begin{subfigure}{.45\linewidth}
        \includegraphics[width=\linewidth]{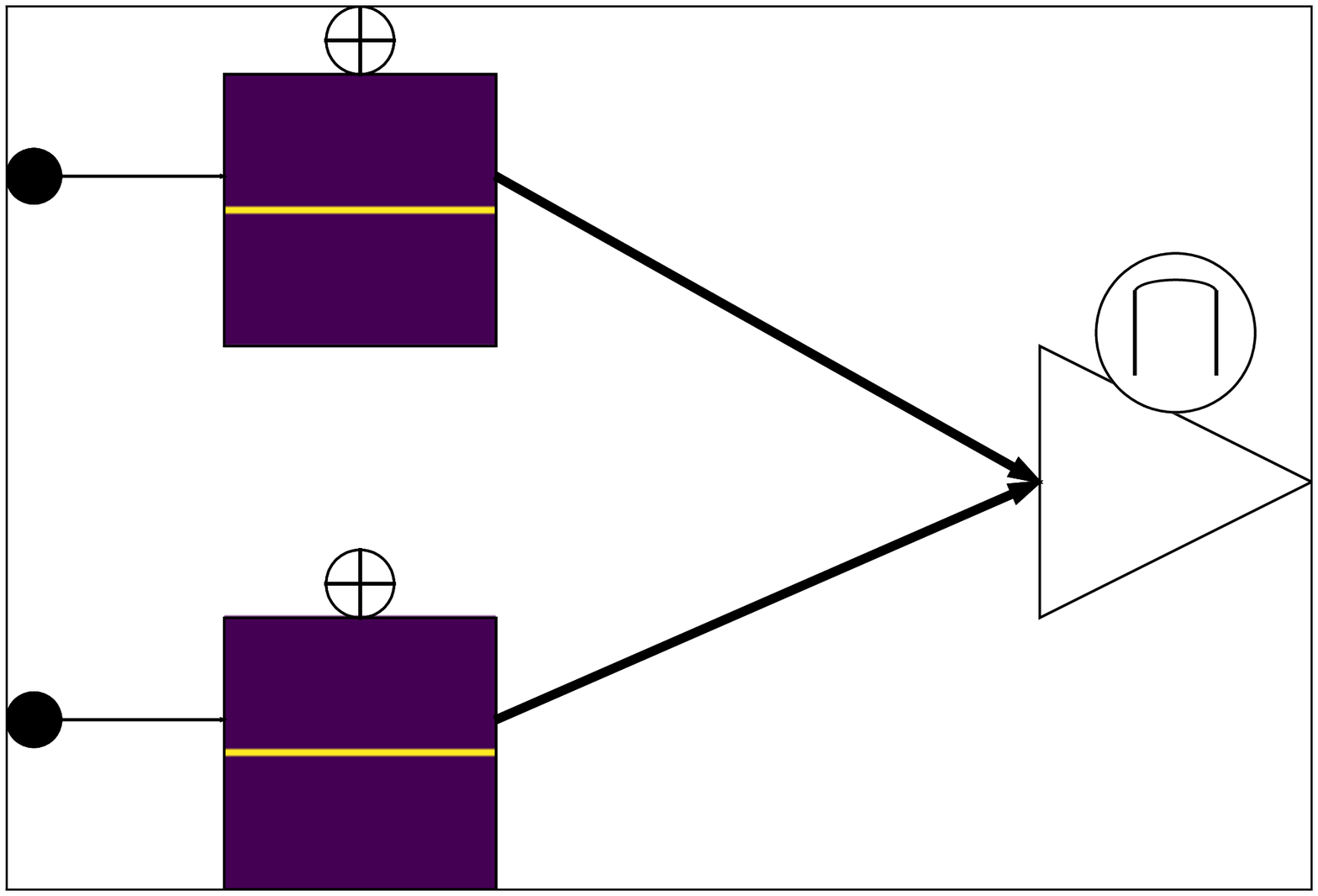}
        \caption{Binarization for architecture 1}
    \end{subfigure}
    \begin{subfigure}{.45\linewidth}
        \includegraphics[width=\linewidth]{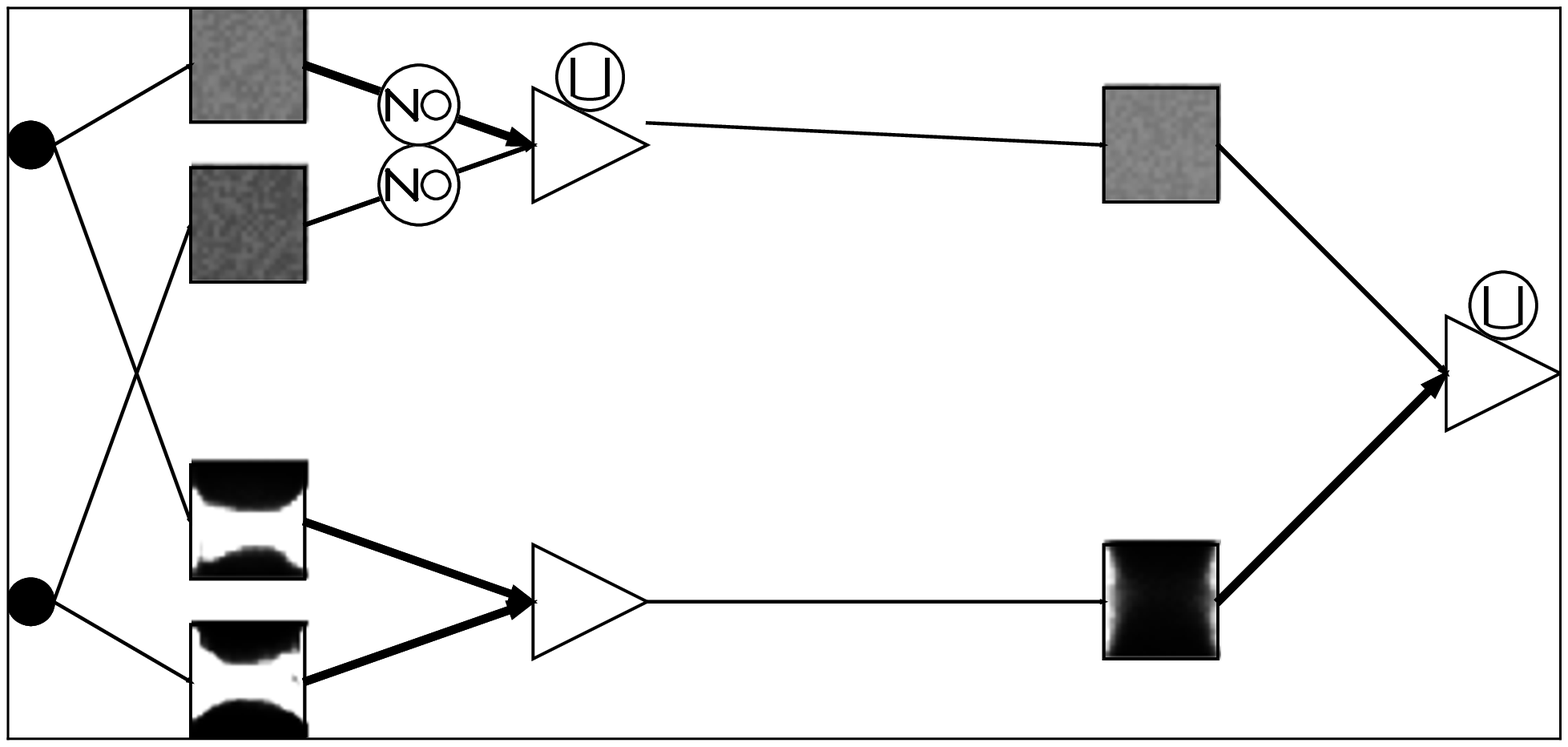}
        \caption{Weights for architecture 2}
    \end{subfigure}
    \begin{subfigure}{.45\linewidth}
        \includegraphics[width=\linewidth]{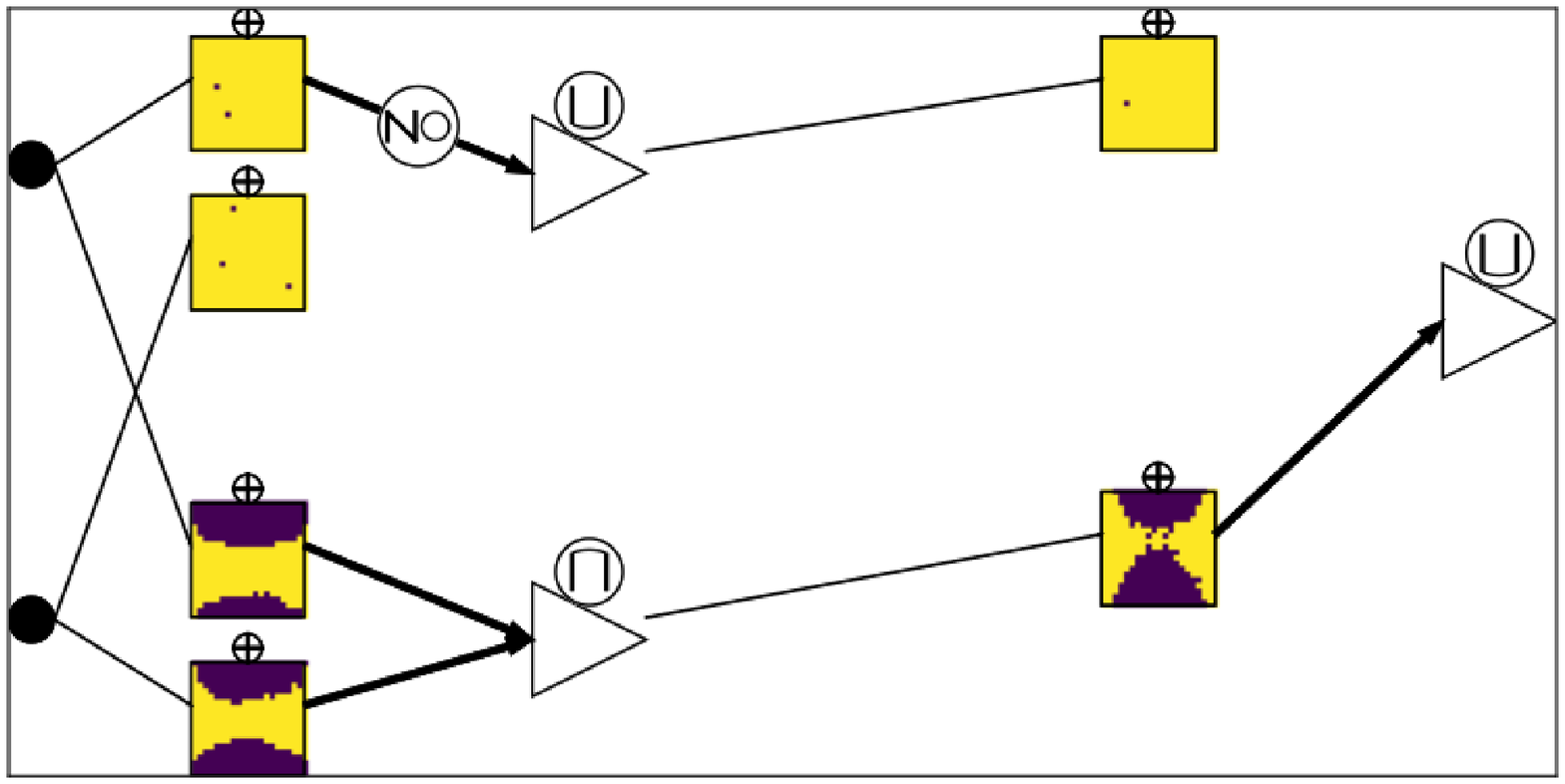}
        \caption{Binarization for architecture 2}
    \end{subfigure}
    \caption{Resulting Sels on axSpA results}
    \label{fig:axspa-results}
\end{figure}

\begin{table}[h]
    \centering
    \caption{Results on opening and closing. DICE error ($1 - \text{DICE}$) is presented for each case: $\R$ before binarization, $\mathbb{B}$ after binarization. \textcolor{blue}{\checkmark} indicates that the neuron is activated.}
    \resizebox{\linewidth}{!}{\newcommand{\opecloselt}[6]{
    \begin{minipage}{.12\linewidth}
        \ifstrequal{#3}{ok}{
          \center{\textcolor{blue}{\checkmark}}
          }{
            \center{\textcolor{red}{$\times$}}
        }
          \includegraphics[width=\textwidth]{#1}
        
         \ifstrequal{#4}{ok}{
          \center{\textcolor{blue}{\checkmark}}
          }{
            \center{\textcolor{red}{$\times$}}
        }
          \includegraphics[width=\textwidth]{#2}
          $\R ~~ #5$
          $\mathbb{B} ~~ #6$
    \end{minipage}
}        

          \begin{tabular}{ | c | c | c | c | c | c | c | c | c | c |}
        \hline
        Dataset & \multicolumn{3}{c|}{Diskorect}  & \multicolumn{3}{c|}{MNIST} & \multicolumn{3}{c|}{Inverted MNIST} \\
        \hline
        Operation & Disk & Stick & Cross & Disk & Stick & Cross & Disk & Stick & Cross \\
        \hline
        Target
        & \begin{minipage}{.12\linewidth}
          \includegraphics[width=\textwidth]{selem_results/true_disk7.eps}
        \end{minipage}
        & \begin{minipage}{.12\linewidth}
          \includegraphics[width=\textwidth]{selem_results/true_hstick7.eps}
        \end{minipage}
        & \begin{minipage}{.12\linewidth}
          \includegraphics[width=\textwidth]{selem_results/true_dcross7.eps}
        \end{minipage}
        & \begin{minipage}{.12\linewidth}
          \includegraphics[width=\textwidth]{selem_results/true_disk7.eps}
        \end{minipage}
        & \begin{minipage}{.12\linewidth}
          \includegraphics[width=\textwidth]{selem_results/true_hstick7.eps}
        \end{minipage}
        & \begin{minipage}{.12\linewidth}
          \includegraphics[width=\textwidth]{selem_results/true_dcross7.eps}
        \end{minipage}
        & \begin{minipage}{.12\linewidth}
          \includegraphics[width=\textwidth]{selem_results/true_disk7.eps}
        \end{minipage}
        & \begin{minipage}{.12\linewidth}
          \includegraphics[width=\textwidth]{selem_results/true_hstick7.eps}
        \end{minipage}
        & \begin{minipage}{.12\linewidth}
          \includegraphics[width=\textwidth]{selem_results/true_dcross7.eps}
        \end{minipage}
        \\
        \hline
        Opening $\ope{}{}$
        & 
        \opecloselt{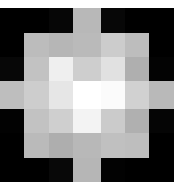}{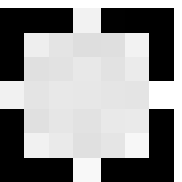}{ko}{ok}{0.000}{0.000}
        & 
        \opecloselt{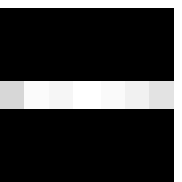}{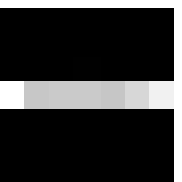}{ok}{ok}{0.000}{0.000}
        & 
        \opecloselt{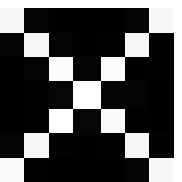}{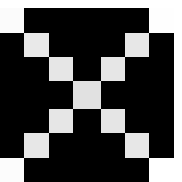}{ko}{ok}{0.000}{0.000}
        & 
        \opecloselt{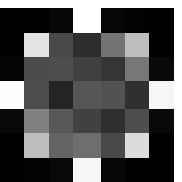}{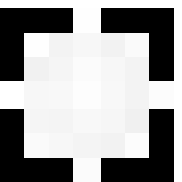}{ko}{ok}{0.000}{0.000}
        & 
        \opecloselt{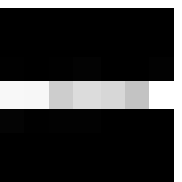}{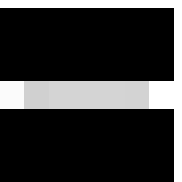}{ko}{ok}{0.000}{0.007}
        & 
        \opecloselt{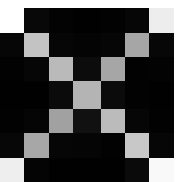}{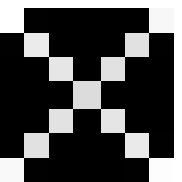}{ko}{ok}{0.000}{0.012}
        & 
        \opecloselt{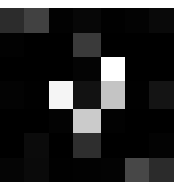}{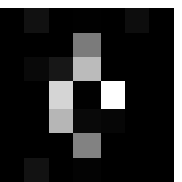}{ko}{ko}{0.006}{0.066}
        & 
        \opecloselt{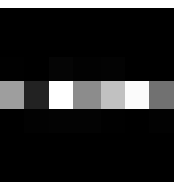}{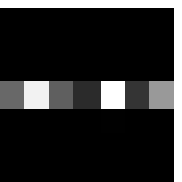}{ko}{ko}{0.001}{0.018}
        & 
        \opecloselt{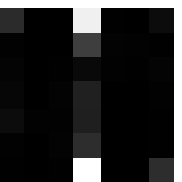}{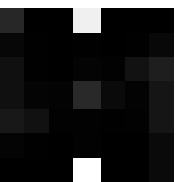}{ko}{ko}{0.014}{0.096}
        \\
        \hline
        Closing $\clos{}{}$
        & 
        \opecloselt{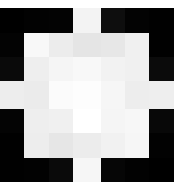}{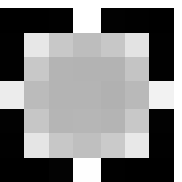}{ok}{ok}{0.000}{0.000}
        & 
        \opecloselt{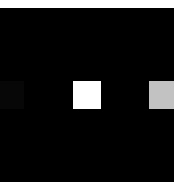}{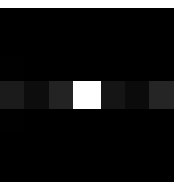}{ok}{ko}{0.077}{0.128}
        & 
        \opecloselt{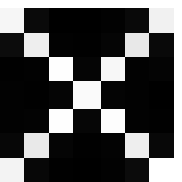}{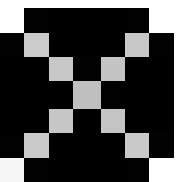}{ko}{ok}{0.000}{0.000}
        & 
        \opecloselt{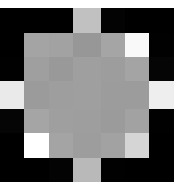}{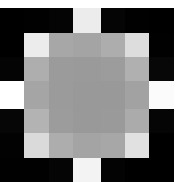}{ok}{ok}{0.000}{0.000}
        & 
        \opecloselt{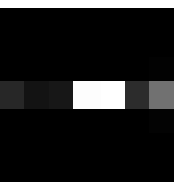}{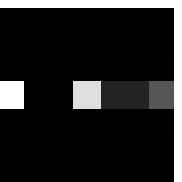}{ko}{ok}{0.000}{0.336}
        & 
        \opecloselt{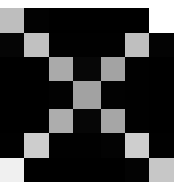}{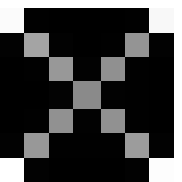}{ko}{ok}{0.000}{0.000}
        & 
        \opecloselt{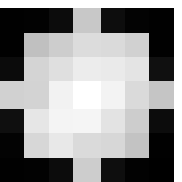}{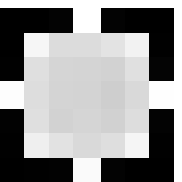}{ok}{ok}{0.000}{0.000}
        & 
        \opecloselt{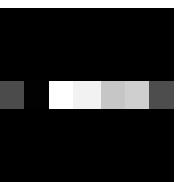}{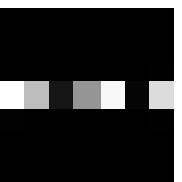}{ok}{ko}{0.000}{0.092}
        &
        \opecloselt{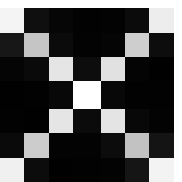}{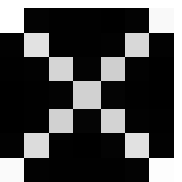}{ko}{ko}{0.000}{0.001}
        \\
        \hline
      \end{tabular}
}
    \label{tab:opeclos}
\end{table}

\end{document}